\def\year{2016}
\documentclass[letterpaper]{article}
\usepackage{aaai16}
\usepackage{times}
\usepackage{helvet}
\usepackage{courier}
\usepackage{setspace}
\usepackage{cprotect}
\usepackage{verbatim}
\usepackage{epsfig}
\usepackage{graphicx}
\usepackage{amsmath}
\usepackage{makeidx}
\usepackage{amssymb}
\usepackage{multirow}
\usepackage{booktabs}

\newcommand{\argmax}{\arg\!\max}

\begin{document}
	\title{Toward a Taxonomy and Computational Models of Abnormalities in Images}
	
	\author{\normalsize Babak Saleh \thanks{Corresponding author: babaks@cs.rutgers.edu}\\
		\small Dept. of Computer Science\\ 
		\small Rutgers University\\
		\small New Jersey , USA
		\And
		\normalsize Ahmed Elgammal\\
		\small Dept. of Computer Science \\ 
		\small Rutgers University\\
		\small New Jersey , USA
		\And
		\normalsize Jacob Feldman \\
		\small Center for Cognitive Science, \\ 
		\small Dept. of Psychology \\ 
		\small Rutgers University\\
		\small New Jersey , USA
		\And
		\normalsize Ali Farhadi \\
		\small Allen Institute for AI, ~\&\\ 
		\small Dept. of Computer Science\\ 
		\small University of Washington\\
		\small Washington , USA}
		\maketitle
	\begin{abstract}
		\begin{quote}
			The human visual system can spot an abnormal image, and reason about what makes it strange. This task has not received enough attention in computer vision. In this paper we study various types of atypicalities in images in a more comprehensive way than has been done before. We propose a new dataset of abnormal images showing a wide range of atypicalities. We design human subject experiments to discover a coarse taxonomy of the reasons for abnormality. Our experiments reveal three major categories of abnormality: object-centric, scene-centric, and contextual. Based on this taxonomy, we propose a comprehensive computational model that can predict all different types of abnormality in images and outperform prior arts in abnormality recognition.

		\end{quote}
	\end{abstract}
	\vspace*{-10pt}
\section{Introduction}
\label{intro}
Humans begin to form categories and abstractions at an early age~\cite{bigbook}. The mechanisms underlying human category formation are the subject of many competing accounts, including those based on prototypes~\cite{minda}, exemplars~\cite{nosofsky}, density estimation~\cite{Ashbydensity}, and Bayesian inference~\cite{goodman}. But all modern models agree that human category representations involve subjective variations in the typicality or probability of objects within categories. For example, bird category includes both highly typical examples such as robins, as well as extremely atypical examples like penguins and ostriches, which while belonging to the category seem like subjectively ``abnormal" examples.  Visual images can seem abnormal, in that they can exhibit features that depart in some way from what is typical for the categories to which they belong. In this paper, we ask what makes visual images seem abnormal with respect to their apparent categories, something that human observers can readily judge but is difficult to capture computationally.

\begin{figure}[t]
	\includegraphics[width=1.08\linewidth,height=.68\linewidth]{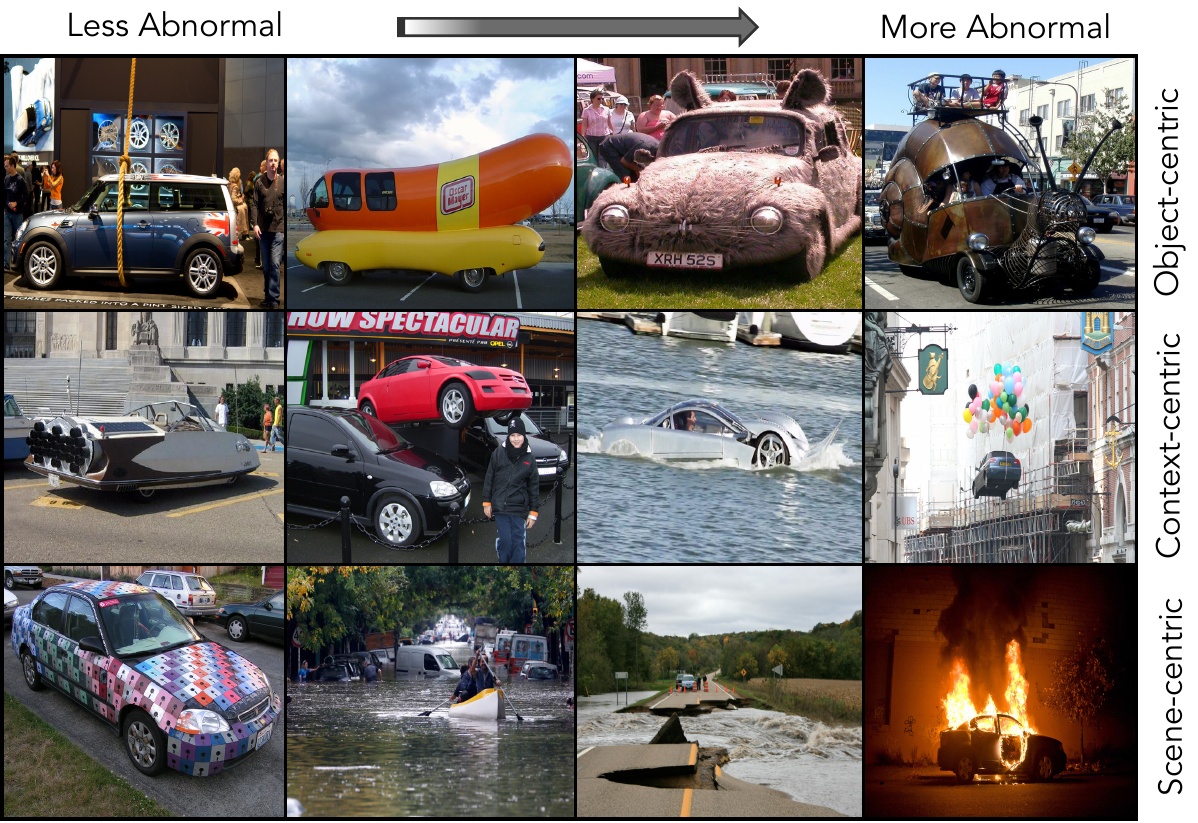}
	\caption{We conducted human subject experiments and found a collection of abnormality reasons that can be clustered into three main categories: Object-centric, Context-centric and Scene-centric. In each row of this figure, we focus on one of these categories and sort abnormal images of cars (from a novel dataset that we propose) based on their surprise scores. These scores are the output of the computational models that we build based on the adopted taxonomy.}
	\vspace{-12pt}
	\label{fig:intro}
\end{figure}

Abnormality detection plays a substantial role in broad range of tasks: learning visual concepts~\cite{Saleh2013}, natural languages processing~\cite{word}, human perception and cognition~\cite{biederman1987}, human action recognition~\cite{videoanomaly}, etc. In computer vision, abnormality in the context of object and scene categorization is an under-studied problem. In contrast to humans ability to generalize and successfully categorize atypical instances~\cite{Rosch75,keselman2005generic}, state-of-the-art computer vision algorithms fail to achieve similar generalization. Table~\ref{tab:deeplearning} shows categorization results of several state-of-the-art approaches~\cite{krizhevsky2012imagenet,sermanet-iclr-14,jia2014caffe,simonyan2014very} when tested on our dataset of abnormal images. This highlights the lack of generalization of these approaches and the importance of studying the problem of abnormality. In this paper we focus on understanding the causes of abnormality and reasoning about it. We address the problem of abnormalities in images and present computational models to detect atypical instances of categories, where the model is trained only on normal images.

\smallskip
\noindent{\bf Challenges:}
There are several issues and concerns in abnormality detection. \textit{First}, researchers are not in an agreement about what is a typical sample of a category and what makes humans distinguish typical instances from atypical ones~\cite{rosch2011slow}. The definition of abnormality in the visual space is even more complex. For example, there is no general rule as what is a typical car. Even if there were such a rule, it might vary across people and categories. 

\textit{Second}, abnormality (atypicality)\footnote{ We will use typicality/atypicality when referring to objects, scenes and context, while we will use normality/abnormality when referring to images. However, at some points we use these words interchangeably.} in images is a complex notion that happens because of a diverse set of reasons that can be related to shape, texture, color, context, pose, location or even a combination of them. Figure~\ref{fig:intro} shows large variability of abnormality reasons among examples of car images that human subjects denoted as abnormal. 

\textit{Third}, there is a gradual transition from typical to atypical instances, so simple discriminative boundary learning between typical and atypical instances does not seem appropriate. Fourth, with the limited number of abnormal images it is hard to be comprehensive with all aspects of abnormality.  This suggests that computational models for identifying abnormalities should not rely on training samples of abnormal images. This is also aligned with how humans are capable of recognizing abnormal images while only observing typical samples~\cite{Rosch75}.

{\em The goal of this paper is to extract a list of reasons of atypicality, enumerating distinct modes or types of abnormal images, and to derive computational models motivated by human abnormality classification.} The contribution of this paper is manifold. We conduct a human-subject experiment to determine a typology of images judged abnormal by human observers and collect data that facilitates discovery of a taxonomy of atypicality. Analysis of the data lead  us to a coarse taxonomy of three reasons for abnormality: object-centric, scene-centric, and contextual.  

We propose a  model for normality (typicality) in images and find meaningful deviations from this normality model as the case of abnormality. The model is fully trained using only normal images. Based on the model we propose and evaluate scoring functions that can predict the three types of atypicalities.  In contrast to prior work on this topic, which were limited to specific types of abnormality (as will be discussed in the next section) our work is a more comprehensive study that is supported and justified by perceptual experiments. We also propose a new dataset for image abnormality research that is more complete than what was previously used in this area. With the final version of this paper we will publish the dataset along with the human subject experiment results and implementation of our computational models.
	
\section{Related Work}
\label{sec:rel}
Space does not allow an encyclopedic review of the anomaly detection literature [See~\cite{survey}]. In computer vision, researchers have studied abnormality detection in events or activities in videos as deviation from normal temporal patterns~\cite{zhong}; modeling out-of-context objects using a tree model that encodes normal scene layout ~\cite{Choi2012}; identifying abnormal images using contextual cues such as unexpected relative location or size of objects~\cite{Park2012}; reasoning about object-centric abnormalities using an attribute-based framework~\cite{Saleh2013}; or investigating the connection between image memorability and abnormality~\cite{isola,ICCV15_Khosla}. From a cognitive point of view, abnormality has also been studied~\cite{josh,rosch1976}; or by a combination of human intervention and scene classifiers on SUN dataset~\cite{Ehinger2011}. Biederman~(\citeyear{biederman1981semantics}) investigates violations in a scene according to five semantic relations: support, interposition, probability, position and size.

\begin{table}
	\begin{center}
		\resizebox{.5\textwidth}{!}{
			\begin{tabular}{|l|l|l|}
				\hline
				Method & Top-1 error ($\%$) & Top-5 error ($\%$) \\
				\hline \hline
				\cite{krizhevsky2012imagenet}  & 74.96 (38.1) & 47.07 (15.32)\\
				\cite{sermanet-iclr-14}& 75.62 (35.1) & 46.73 (14.2)\\
				\cite{jia2014caffe}& 77.12 (39.4) & 46.86 (16.6)\\
				VGG-16~\cite{simonyan2014very} & 77.82 (30.9) & 47.49 (15.3) \\
				VGG-19~\cite{simonyan2014very} & 76.35 (30.5) & 45.99 (15.2) \\
				\hline
			\end{tabular}}
		\end{center}
		\vspace*{-5pt}
		\caption{State-of-the-art Convolutional Neural Networks (trained on normal images) fail to generalize to abnormal images for the task of object classification. Numbers in parenthesis show the reported errors on normal images (ILSVRC 2012 validation data), while numbers next to them is the error on our abnormal images.}
		\vspace*{-10pt}
		\label{tab:deeplearning}
	\end{table}

	Among all possible reasons for abnormality, previous work has mainly focused on isolated and specific kinds of abnormalities; for example, rooted in either objects~\cite{Saleh2013} or context~\cite{Park2012}. In this paper we first investigate the different reasons for abnormality, and provide evidence that supports our proposed clusters of reasons. Following this grouping, we introduce a comprehensive framework for detecting and classifying different types of atypicalities in images.
	
	\begin{figure*}[t]
		\includegraphics[width=\linewidth, height = .33\linewidth]{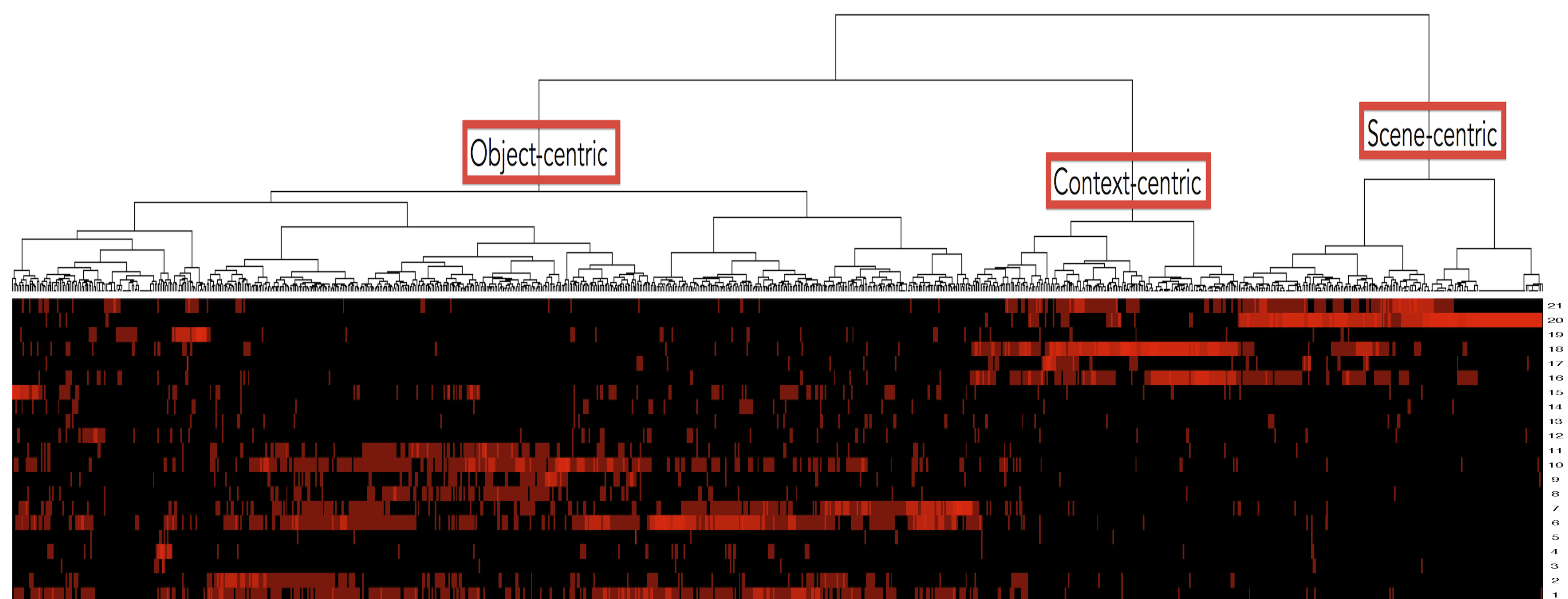}
		\caption{Agglomerative clustering of abnormal images (columns) based on the human subject responses for each abnormality reason (rows). The dendrogram on top of the figure shows how the abnormal images in our dataset can be grouped to make three clusters, reflecting three major latent categories of abnormality. Each cluster corresponds to a specific list of abnormality reasons. Details of these three categories of abnormality can be found in Table~\ref{tab:reason}.}
		\label{fig:turk}
		\vspace{-10pt}
	\end{figure*}

	\section{Taxonomy of Abnormality}
	\label{turk}
	The taxonomy of abnormality in images is not well-defined either in Psychology or Computer Vision. We design a human subject experiment to discover a coarse taxonomy for abnormality. To this end, we first need to collect a dataset of abnormal images that is more comprehensive than what has been used in prior work. 

	\medskip
	\noindent{\bf 1001 Abnormal Image Dataset:}
	Previous datasets for abnormality research are either not specifically designed for the task of abnormality detection~\cite{Choi2012}, or limited to a specific type of abnormality [\cite{Choi2012,Park2012} are focused on contextual cues, and~\cite{Saleh2013} is concerned with object-centric reasons)]; or of small size [\cite{Choi2012} has 40 and~\cite{Park2012} has 200 images].
	
	In order to study abnormalities in images with more details, this paper introduces a dataset that is more comprehensive both in terms of the number of images and types of abnormality. To collect this dataset, we started by gathering images from three public datasets used in~\cite{Choi2012,Park2012,Saleh2013}, which we call ``initial collection" and almost doubled the size by adding more images from the web. Our image collection process is similar to~\cite{Saleh2013}, but textual queries that we used for image search are not limited to abnormal objects. For examples, we used ``strange street" or ``weird living room" as additional queries. After downloading a large number of images, we pruned the result by removing duplicates and very low-quality images. Then we merged these images and ``initial collection" into the final dataset with a total number of 1001 unique abnormal images. Figure~\ref{fig:intro} shows some images of this dataset. We validated our collection and acquired image annotations by conducting a  human-subject experiment, which is explained next.
	
	\begin{center}
		\begin{table*}
			\begin{tabular}{p{0.15\textwidth}||p{0.8\textwidth}}
				\hline
				Main Category & Detailed Reasons in Amazon Mechanical Turk Experiment	 \\ \hline\hline
				Scene-centric & Strange event happening in the scene(21); Strange scene(20)\\ \hline
				Context-centric &  Atypical object size(19); Strange location of the object(18); Atypical object pose(17); Weird combination of objects and scene(16)\\ \hline
				Object-centric & Unexpected part(7); Weird shaped part(6); Misplaced part(5); Missing part(4); Body posture(14); Mixture of object classes(13); Un-nameable shape(12); Object is not complete(3); Unknown object(15); Object in the shape of another object(2); Atypical pattern(10), Weird color(9), Strange material(11), Weird texture(8); Strange object contour(1)\\
				\hline
			\end{tabular}
			\caption{Learned taxonomy for reasons of abnormality in images based on our human subject experiment. Numbers in the parenthesis are indexes of reasons, which correspond to the rows in Figure~\ref{fig:turk}.}
			\label{tab:reason}
			\vspace*{-10pt}
		\end{table*}
	\end{center}
	
	\vspace{-28pt}
	\paragraph{Human Subject Experiment:}
	We conducted a two-phase experiment. First, we asked four subjects to take an on-site test and exposed each subject to a unique set of images from our dataset. The human subject was asked to determine whether the images are abnormal, and if they are, to explain the reason behind the abnormality in their own words. The goal of this step is to compile an initial comprehensive list of reasons for abnormality in images. 
	
	We enumerated the responses into a list, and did not merge them unless two reasons clearly refer to the same notion (e.g. ``This object does not have an expected part" and ``One part is missing for this object" are classified as the same reason). By this process we came up with a list of 21 fine-grained reasons for abnormality written in plain English. Some example reasons include ``An unexpected event is happening in this image", `` Weird object material" and ``Missing part for the object". The full list of fine-grained reasons is shown in Table~\ref{tab:reason}. We denote this list by ``expanded abnormality list''. We understand that this list might not be universal for all possible reasons of visual abnormality, but we believe it covers most types of abnormalities in our dataset. 
	
	In the second phase, our goal was to annotate all images in our dataset with a reasonable number of human subject responses, and discover a hierarchy of these reasons via an unsupervised approach. In order to complete this large-scale experiment, we asked annotators on Amazon Mechanical Turk to annotate images in our dataset based on the 21 reasons in the expanded abnormality list. Also we added two extra choices: \textit{``This image looks abnormal to me, but I cannot name the reason"} or \textit{``Abnormal for a reason that is not listed"} followed by a text box that the subject could write in it. This gives annotators the opportunity of describing the abnormality in their own words. 
	
	As one image can look abnormal because of multiple cues, annotators could select multiple reasons and were not limited to the 21 reasons on the list. We picked annotators with a good history for the task of image annotation and categorization (users for whom at least 95\% of previous responses over the past three months were accepted). Subjects could not take a task twice and  for each image we aggregated responses from six unique human subjects. To verify the quality of annotations, we randomly asked annotators to take an image for the second time to see if their response matched his/her previous response. Due to the importance of non-random responses, if an annotator showed a random behavior in choosing the reasons of abnormality, we re-sent the task to the rest of participants and stopped the suspicious annotator from taking future HITs. In total 60 unique human subjects participated in this experiment.
	\vspace{-10pt}
	\paragraph{Discovering a Taxonomy of Abnormality:}
	We averaged the responses across all subjects for every image and for each of the 21 reasons. This results in a an embedding of the images into a 21-dimensional space, i.e. each image is represented with a 21-dimensional response vector. We hypothesize that there is a latent abnormality subspace (space of reasons for abnormality in images); and measuring similarity between the response vectors for images is expected to reflect the similarity between them in the latent abnormality space. To discover a taxonomy of abnormality, we performed unsupervised learning on the collection of response vectors using bottom-up agglomerative clustering. We used the Euclidean distance as a dissimilarity measure and the Ward's minimum variance criteria~\cite{ward} for the linkage function. At each step, a pair of clusters that result in the minimum increase in the within-cluster variance are merged.
	
	Figure~\ref{fig:turk} shows the resulted dendrogram of images and the corresponding responses in the 21-dimensional space. One can spot three main clusters in this dendrogram, which directly corresponds to grouping of reasons of abnormality. The implied grouping is shown in Table~\ref{tab:reason}. Consequently, we can name intuitive atypicality groups based on this coarse taxonomy: Scene-centric atypicality, Context-centric atypicality,  and Object-centric atypicality.  We performed several experiments on clustering with different linkage functions and metrics; however, we observed that this coarse taxonomy is robust over changes in the clustering parameters. It is interesting that prior research is broadly consistent with this taxonomy: the work of~\cite{Park2012,Choi2012,Ehinger2011} proposed models to predict contextual atypicality, and proposed models of~\cite{Saleh2013} predict object-centric abnormality. Thus we conclude that our taxonomy, which is motivated by human judgments, encompasses previous approaches in a more systematic way.
	
	\section{Computational Framework}
	\label{sec:model}
In this section, we propose a computational model to find abnormal images and reason about them based on three scores that come from three main groups of abnormality. We start by investigating normal images and proposing a model for relating elements of an image: object, context and scene. Next, we derive a set of scores, called ``Surprise Scores", to measure how abnormal  an image is with respect to these elements. Later we explain how we merge the different scores to decide if the image is abnormal or not, and finally find the dominating abnormality reason that affects this decision. 

\subsection{Modeling Typicality}
\begin{figure}[t] 
	\begin{center}
		\includegraphics[width=.33\textwidth,height=.18\textwidth]{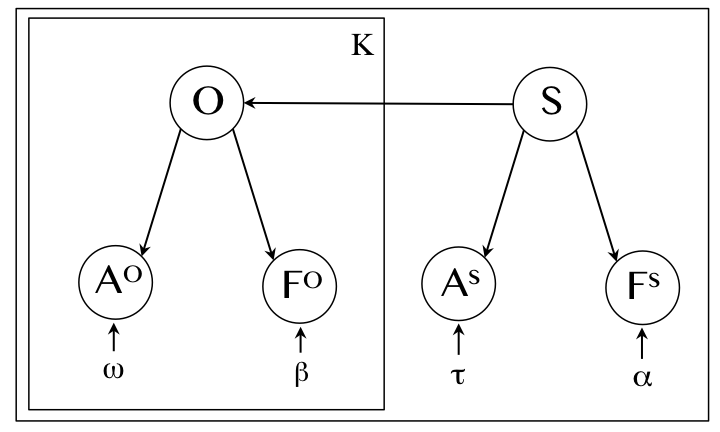}
	\end{center}
	\vspace*{-15pt}
	\caption{Graphical Model of Normal Images}
	\label{fig:model} 
\end{figure}

We propose a Bayesian generative model for typical scenes and objects, depicted in Figure~\ref{fig:model}. This model formulates the relation between objects, context and other information in the scene that is not captured by objects or the context (e.g. scene characteristics such as Sunny or Crowded).  This is a model of typicality, and atypicality/abnormality is detected as a deviation from typicality. Hence, this model is trained using only typical images and relies on visual attributes and categories of both objects and scenes. 

Visual attributes have been studied extensively in recognition~\cite{lad2014interactively,parikh2011relative}. In contrast to low-level visual features (e.g. HOG, SIFT), attributes represent a valuable intermediate semantic representation of images that are human understandable (nameable). Example attributes can be ``Open area", ``Sunny weather" for scenes and ``wooden'' or ``spotty" for objects.
Attributes are powerful tools for judging about abnormality. For example, the object-centric model of~\cite{Saleh2013} mainly used attribute classifiers to reason about abnormality. However, the response of an attribute classifier is noisy and uncertain. As a result, we categorize the object based on low-level visual features apart from its attributes scores. Later, out model at the level of the object focuses on deviations between categories of the objects and its meaningful visual characteristics (attributes). In short, if low-level features predict an object to be a car, while attribute responses do not provide evidence for a car, that is an indication of abnormality.

As a similar argument stands at the level of scenes, we model the typicality of low-level visual features ($F$) and attributes ($A$) for both objects ($O$) and scenes ($S$). Figure~\ref{fig:model} shows that assuming we observe a normal image~\textit{I}, any distribution over scene category $S$ imposes a distribution over the categories of objects $O$ that are present. This procedure holds for all~\textit{$K$} objects in the image (left plate is repeated $K$ times). Each object category imposes a distribution over object's low-level features $F^o$ and attributes $A^o$. Similarly, scene categories impose a distribution over scene's low-level features $F^s$ and attributes $A^s$. However, extracted visual features for scenes are different from ones extracted for objects. We define two disjoint sets of attributes for objects ($A^o = \{ A^o_i\}_1^n $) and attributes for scenes ($A^s = \{ A^s_i\}_1^m $).

Learning the model involves learning the conditional distribution of object-attribute, given object categories ($ \{P(A^o_i | O_k) , i=1\cdots n, k=1\cdots V \} $),  and scene-attribute conditional probability distribution given scene categories ( $ \{P(A^s_i | S_j), i=1\cdots m, j=1\cdots J  \}$), where each of these distributions is modeled as a Gaussian. We also learn probabilities of object categories given scene categories ($ \{P(O_k|S_j), k=1\cdots V, j=1\cdots J  \}$) , where $V$ and $J$ are number of object and scene categories.

\subsection{Measuring Abnormality of Images}
\paragraph{Scene-centric Abnormality Score:}
For any scene category, some visual attributes are more relevant (expected). This is what we call relevance of $i^{th}$ scene attribute for the $j^{th}$ scene category, denoted by $\Omega(A^{s}_{i}, S_{j})$~\footnote{For simplicity, we slightly abuse the notation and use $A^{s}_i$ to denote both the $i^{th}$ attribute, and the $i^{th}$ attribute classifier response for scene attributes. The same holds for object attributes as well.}. We compute this term by calculating the reciprocal of the entropy of the scene-level attributes for a given scene category $\Omega(A^{s}_{i}, S_{j})=1/H(A^{s}_i|S_{j})$ over normal images. This relevance term does not depend on the test image.

For a given image, applying scene classifiers produce a distribution over scene categories. Assuming a scene category, we compute the information content in each scene-attribute classifier response ($I(A^s_i|S_{j}) = -log P(A^s_{i}|S_{j})$). This information content is a measure of the surprise by observing an attribute for a given scene class. Since attribute classifiers are noisy, depending on the concept that they are modeling,  we need to attenuate the surprise score of a given attribute by how accurate is the attribute classifier. We denote this term by $\Upsilon(A^s_{i})$, which measures the accuracy of the $i^{th}$ scene attribute classifier on normal images. Therefore the scene surprise score (\textit{Surprise$_{S}$}) is computed by taking the expectation given $P(S_j)$ as following:

\begin{equation}
	\sum_j P(S_j) 
	[\sum_{i} I(A^{s}_{i}|S_{j})  \Upsilon(A^{s}_{i})  \Omega(A^{s}_{i}, S_{j})]
	\label{eq:scene}
\end{equation}

\vspace{-20pt}
\paragraph{Context-centric  Abnormality Score:}
An image looks abnormal due to its atypical context if one of the following happens: first, an unexpected occurrence of object(s) in a given scene. (e.g. elephant in the room); second, strange locations of objects in the scene (e.g. a car on top of the house); or inappropriate relative size of the object. We propose Eq.~\ref{eq:context-all} to measure the context-centric surprise (Surprise$_{C}$) of an image based on aforementioned reasons: 

\begin{equation}
	\sum_{k} \sum_{j} \Lambda(O_k)  [\hat{I}(O_k | S_j) + I(L_k | O_k)].
	\label{eq:context-all}
\end{equation}

The term $\hat{I}(O, S)$ measures the amount of surprise stemming form the co-occurrence of the objects in the scene (Eq.~\ref{eq:context}). 
We measure the surprise associated with each object classes appearing in the scene $S_{j}$  by computing the information content of each combination of scene categories and object classes, $I(O_{k}|S_{j})$, modulated by the probability of the object and scene categories.
\begin{equation}
	\hat{I}(O_k | S_j) =  P(S_{j}) P(O_{k}) I(O_{k}|S_{j}).
	\label{eq:context}
\end{equation}
On the grounds that we use a distribution as the output of classifiers rather than a single class confidence, we do not need to involve the accuracy of neither the object classifier nor the scene classifier to tackle the uncertainty output.

The term  $I(L_k | O_k)$  measures how much surprising is the location of the object $k$ in the image. Assuming we know category of the object ($O_k$), we expect to see it in certain locations in the image. By considering one object category at a time, we learn a distribution of possible locations for the object in normal images and use it to compute the information content of the object location in a test image.

Finally we aggregate the co-occurrence and location term and modulate  the score by multiplying it with $\Lambda(O_k)$, which stands for the importance of the size of the object relative to the whole image in judging the context atypicality. If the object of interest is tiny or huge in the image, the contextual surprise should be modulated down. To model $\Lambda(O)$ for each object category($O$) we learn the distribution of its relative size by considering the normal images with typical context and for the test image compute its probability based on this distribution. 
\vspace{-5pt}
\paragraph{Object-centric Abnormality:}
For \textit{Surprise$_O$} we check if the objects in the image look typical or not independently. We assume that we take the object out of the scene and measure how abnormal it is based on its predicted object class and visual attributes. This term is in part similar to work of Saleh et al~\cite{Saleh2013}. However, we are different from their work as we classify the objects  based on low-level visual features $F^o$ rather than visual attributes $A^o$. We formulate the object-centric surprise score (Surprise$_O$) as:

\begin{equation}
	\sum_{k}P(O_k) *  (\sum_{i}  I(A^o_{i} |O_k) * \Upsilon(A^o_{i}) * \Omega(A^o_{i},O_k))
	\label{eq:object}
\end{equation}

Where $P(O_k)$ is the distribution over object categories obtained from low-level visual features. $I(A^o_{i} |O_k) = -log(P(A^o_{i} |O_k))$ denotes the amount of the surprise by observing the response of the i-th attribute classifier, given class $O_k$. Similar to scene-centric surprise score, $\Upsilon(A^o_{i})$ adjusts the weights of visual attributes based on how reliable one attribute performs on normal images. $\Omega(A^o_{i},O_k)$ models the relevance of attribute $A^o_{i}$ to object $k$, however this is computed based on ground truth annotation rather than the conditional entropy of attributes.

\subsection{Parametric Model for Typicality}
For the final decision about abnormality of an image we should compare the three surprise scores and pick the maximum as the the most important reason of abnormality. However, there are two issues that prevent us from using the maximum of raw surprise scores. 
These described surprise scores are based on quantifying the information content, therefore these measures are unbounded (as the probability approaches zero, the surprise approaches infinity). The other issue is that these surprise scores are not comparable since the information content in each of them are modulated differently. As a result it is hard to compare the values of \textit{Surprise$_O$}, \textit{Surprise$_S$}, and \textit{Surprise$_C$} to determine which of these reasons gives rise to the abnormality in the image, if any. To tackle these issues, we propose to model the distribution of the surprise scores for normal images.

Toward this goal, we compare fitting different parametric models to the empirical distributions of three surprise scores, computed over normal images. For model selection we consider simplicity of the distribution, as well as how well it fits the empirical data based on Akaike Information Criterion (AIC)~\cite{AIC}. We are interested in simpler distributions, because of their better generalization and the ability to derive simpler (sometime closed from) CDFs. Our experiments show that independent of the reason of abnormality, surprise scores follow exponential family of distributions. We pick \textit{``Inverse Gaussian"} distribution as the underlying distribution. Due to limited space, we put more analysis in the supplementary material. Given these probabilistic models, we can compute the probability of observing a given surprise score instead of the raw surprise scores.
Then we can classify the reason of abnormality in an image by comparing the CDFs of the parametric models, i.e., 
\begin{equation}
	\label{eq:parametric_model}
	\argmax_{o,s,c}  (\phi_o({Surprise_O}),\phi_s({Surprise_S}),\phi_c({Surprise_C}) )
\end{equation}
Where $\phi_o(\cdot)$,$\phi_s(\cdot)$,$\phi_c(\cdot)$ are the inverse Gaussian CDFs for the object, scene, and context -centric parametric surprise models respectively. Parameters of each model are estimated only from the normal training data.

	\section{Experiments and Results}
	\label{exp}
\subsection{Object-centric Typicality Modeling}
We train our model for abnormality prediction on six classes of objects: Airplane, Boat, Car, Chair, Motorbike and Sofa. We choose these categories to be comparable with related work~\cite{Saleh2013}. Based on our experiments state-of-the-art object detectors~\cite{rcnn,endres} generally fail to detect abnormal objects. As a result, we assume that object bounding boxes are given in the image. Through our experiments, we convert confidences of classifiers (e.g. attribute classifiers) to the probability by using Platt's method~\cite{platt}.

\begin{table}[t]
	\begin{center}
		\begin{tabular}{|l|l|l|l|l|}
			\hline
			Reason-name & Var. I & Var. II & Var. III & Full-score\\
			\hline \hline
			Object-centric &  0.6128 & 0.7985 & 0.8050 & 0.8662\\
			Context-centric & 0.6778  & 0.6923 & 0.8255 & 0.8517\\
			Scene-centric & 0.6625  & 0.7133 & 0.7210 & 0.7418\\
			\hline
		\end{tabular}
	\end{center}
	\caption{Ablation experiment to evaluate the importance of different elements of each surprise score (rows) for the task of abnormality classification (Area Under Curve - AUC). For scene-centric and object-centeric: Var.I) Only $I(A|S (or C))$, Var.II) Full-score without relevance ($\Omega(A, S (or C))$), Var.III) Full score without attribute accuracy ($\Upsilon(A)$). For context-centric: Var.I) $I(O|S)$, Var.II) $I(O|S)*\Lambda(O_k)$, Var.III) $I(O|S) + I(L|O)$.}
	\label{tab:ablations}
	\vspace*{-10pt}
\end{table}

\vspace{-5pt}
\paragraph{Object Classification}
We use ``Kernel Descriptors" of Bo et al~\cite{bo_nips10} to extract low-level visual features for each object. We specifically use \textit{Gradient Match Kernels, Color Match Kernel, Local Binary Pattern Match} kernels. We compute these kernel descriptors on fixed size 16 x 16 local image patches, sampled densely over a grid with step size 8 in a spatial pyramid setting with four layers. This results in a 4000 dimensional feature vector. We train a set of one-vs-all SVM classifiers for each object class using normal images in PASCAL train set. We perform five-fold cross validation to find the best values for parameters of the SVM. This achieves in 87.46\% average precision for the task of object classification in PASCAL2010 test set. Object classification in abnormal images is extremely challenging and state-of-the-art approaches cannot generalize to atypical objects (see Table~\ref{tab:deeplearning}). Our learned object classifiers achieve top-1 error 67.25\% on abnormal objects.

\vspace{-5pt}
\paragraph{Object Attributes}
We use the annotation of 64 visual attributes for the objects in ``aPASCAL" dataset~\cite{AliAtr2009}.
Farhadi et al~\cite{AliAtr2009} extracted HOG, color, edges and texture as base features and learned important dimensions of this feature vector for each attribute using $l1$-regularized regression. However, we do not extract edges and we extract colorSIFT~\cite{colorsift} rather than simple color descriptors. Also we do not perform the feature selection and use the original base features. Our approach for learning attribute classifiers outperform pre-trained classifiers of~\cite{AliAtr2009} for the task of attribute prediction on aPascal test set. 

\begin{table*}
	\begin{center}
		\begin{small}
			\resizebox{\textwidth}{!}{
				\begin{tabular}{|c|c|c|c|c|c|c|}
					\hline
					Experiment & Method & Accuracy & \multicolumn{2}{c|}{Training images} & \multicolumn{2}{c|}{Testing images} \\
					\cline{4-7}
					Number & & & Normal & Abnormal & Normal & Abnormal\\
					\hline \hline
					I & Object-centric baseline~\cite{Saleh2013} & 0.9125  & Pascal & Not Used   & Pascal & Dataset of~\cite{Saleh2013}\\
					& Our Model - Object-centric & \textbf{0.9311}       & Pascal & Not Used   & Pascal & Dataset of~\cite{Saleh2013}\\ \hline
					II & Context-centric baseline~\cite{Park2012}& 0.8518   & SUN    & Not Used   & SUN    & Subset of~\cite{Park2012}-without human\\
					& Our Model - Context-centric & \textbf{0.8943}      & Pascal & Not Used   & SUN    & Subset of~\cite{Park2012}-without human\\ \hline
					III & One Class SVM - based on Attributes & 0.5361          & Pascal & Not Used   & Pascal & Our dataset\\
					& Two Class SVM - based on Attributes & 0.7855          & Pascal & Our dataset& Pascal & Our dataset\\
					& One class SVM - based on Deep features (fc6) & 0.5969          & Pascal & Not Used   & Pascal & Our dataset\\
					& Two class SVM - based on Deep features (fc6) & 0.8524          & Pascal & Our dataset& Pascal & Our dataset\\
					\hline
					IV & Our Model - No Object-centric score &  0.8004      & Pascal & Not Used   & Pascal & Our dataset\\
					& Our Model - No Context-centric score &  0.8863     & Pascal & Not Used   & Pascal & Our dataset\\
					& Our Model - No Scene-centric score &  0.8635       & Pascal & Not Used   & Pascal & Our dataset\\
					& Our Model - All three reasons &  \textbf{0.8914}   & Pascal & Not Used   & Pascal & Our dataset\\
					\hline
				\end{tabular}}
			\end{small}
		\end{center}
		\caption{Evaluating the performance (AUC) of different methods for classifying normal images vs. abnormal images.}
		\vspace{-10pt}
		\label{tab:detection}
	\end{table*}

	\subsection{Context-centric Typicality Modeling}
	Following Eq.~\ref{eq:context-all} we compute the amount of information provided by the co-occurrence, location and size of the objects in the scene. For modeling the co-occurrence we use the annotation of SUN dataset and learn the conditional entropy of object categories for each scene category. To learn the typical location of objects in images and their relative size, we use PASCAL context dataset~\cite{mottaghi} that annotated PASCAL images with semantic segmentation. For this purpose we divide PASCAL images into equally-sized grids and for each grid compute the probability of the number of pixels that belongs to each object category. We learn these distributions over all images that are labeled as positive samples of the object category. Our experiments show that the ratio of pixels that contribute to a specific object in a grid, follows an {\it Exponential distribution}. We model the normal relative size (the ratio of object to the whole image) with a {\it Gamma distribution}.

	\subsection{Scene-centric Typicality Modeling}
	To model the typical scene and context, we use the annotation of SUN dataset~\cite{SUN} to find most frequent scene categories for our six object classes. We start with top ten scene categories for each object class and merge them based on  similarities in images,  which results in 4700 images of 16 scene categories. . For example, we merge Airfield, Airport, Runway and Taxiway into one category.
	
	\vspace{-5pt}
	\paragraph*{Scene Classification}
	State-of-the-art for the task of scene classification~\cite{blocks,carl,GigaSUN} use image collections and scene categories that are different from our experimental setting. As a result, we train scene classifiers specifically for our experiments by following the approach of Parizi et al~\cite{parizi2014automatic}. However, we modify the process of selecting image patches during training classifiers. This approach outperforms prior arts for the task of scene categorization of normal images in our collection by achieving 94\% average precision over 16 scene categories in our train set. 
		\vspace{-5pt}
	\paragraph{Scene Attributes}
	We use 102 scene-level visual attributes proposed by Patterson et al~\cite{SunAttr}. We follow the strategy of~\cite{SunAttr} to train attribute classifiers using images of normal scene. We measure the attribute reliability $\Upsilon(A^{S}_i)$ and relevance of an attribute for a scene category, in terms of the conditional entropy of the attribute confidences of all normal images from the same scene category: $H(A_{i}|S_{j})$. We also estimate the conditional distribution of attribute responses in normal images for a given scene category, as a normal distribution and later use this probability in computing $I(A_{i}|S_{j})$ for abnormal images.

	\subsection{Abnormality Classification and Reasoning}
	We compute all three \textit{\textbf{Object-centric, Context-centric}} and \textit{\textbf{Scene-centric}} surprise scores following Eqs.~\ref{eq:scene},\ref{eq:context} \&~\ref{eq:object}. We use these surprise scores to first, classify an image as abnormal vs. normal (abnormality classification). Next, we use the parametric model for abnormality classification and finding the reason of abnormality that contributes the most to our final decision (abnormality reasoning). In the first step, we conduct an ablation experiment to evaluate the performance of each surprise score, and its components for distinguishing normal vs. abnormal images. Table~\ref{tab:ablations} shows the result (AUC) of this experiment, where each row represents a specific reason and columns are different variations of the corresponding surprise score. In each row, we consider the abnormal images of that specific reason as the positive set and all normal images along with other abnormal images (due to a different reason) as the negative set.
	
	Table~\ref{tab:ablations} shows that for all reasons of abnormality, the full version of surprise scores -- all components included -- achieves the best result (last column). For object and scene-centric surprise scores, Var. I represents a variation of the surprise score, which only uses the term $I(A|S (or C))$. We can improve this basic score by adding ``the accuracy of attribute classifiers" (in Var. II), or ``relevance of the attribute to the object/scene category"(in Var. III). We conclude that both components of relevance and attribute accuracy are equally important for improving the performance of abnormality classification. For context-centric surprise scores, the location of the object (conditioned on its category) is by a large margin, the most important factor to improve the basic surprise score (in Var. I) -- which only finds the irregular co-occurrence of objects and scene.

	These reason-specific surprise scores can be used for sorting images based on how abnormal they look like. Figure~\ref{fig:intro} shows some examples of these rankings for images of cars. Each row corresponds to one reason of abnormality, where images of abnormal cars are selected from our dataset and sorted based on the corresponding surprise score. Supplementary material includes more images of ranking experiment, histograms of these individual surprise scores for normal vs. abnormal images and the corresponding fitted probability functions.

	We compute the final surprise score of an image based on the Eq.~\ref{eq:parametric_model}, where we use the index of maximum surprise score for the task of abnormality reasoning. Table~\ref{tab:detection} shows the performance (AUC) of our final model for the task of abnormality classification in four different experiments (four boxes), where the last four columns indicate the source of images that we use for training and testing. Comparing the first two rows show that we outperform the baseline of object-centric abnormality classification~\cite{Saleh2013} on their proposed dataset. This is because we learn better attribute classifiers and compute the surprise score by considering all possible categories for objects. Box II in table~\ref{tab:detection} shows that our proposed context-centric surprise score outperforms state-of-the-art~\cite{Park2012} for contextual abnormality classification. It should be mentioned that Park et al~\cite{Park2012} originally performed the task of abnormal object detection. For the sake of a fair comparison, we change their evaluation methodology to measure their performance for the task of abnormality classification.
	
	Box III in Table~\ref{tab:detection} shows the results of another baseline experiment for abnormality classification, where all abnormal images are used at the test time (despite box I\& II). We train one-class (fifth row) or two-class (sixth row) Support Vector Machines (SVM) classifiers, where the later case performs better. Although we do not use abnormal images in training, our model still outperforms the two-way SVM classifier that is trained via both normal and abnormal images. This is mainly due to the fact that abnormality is a graded phenomena and a generative model finds abnormality better than discriminative ones. To evaluate the importance of each reason-specific surprise score in the parametric model, we conduct an ablation experiment as it is reported in box IV of the Table~\ref{tab:detection}. In each row, we remove one reason of abnormality and compute the parametric model based on the other two surprise scores. Comparing these performances with the one of full model (last row) show that object-centric surprise score is the most important element of the final model, as removing it results in the biggest drop in the performance. Also the context-centric seems to be the least important reason for detecting abnormal images.
	
\begin{figure}[t]
		\includegraphics[width=\linewidth, height =.68\linewidth]{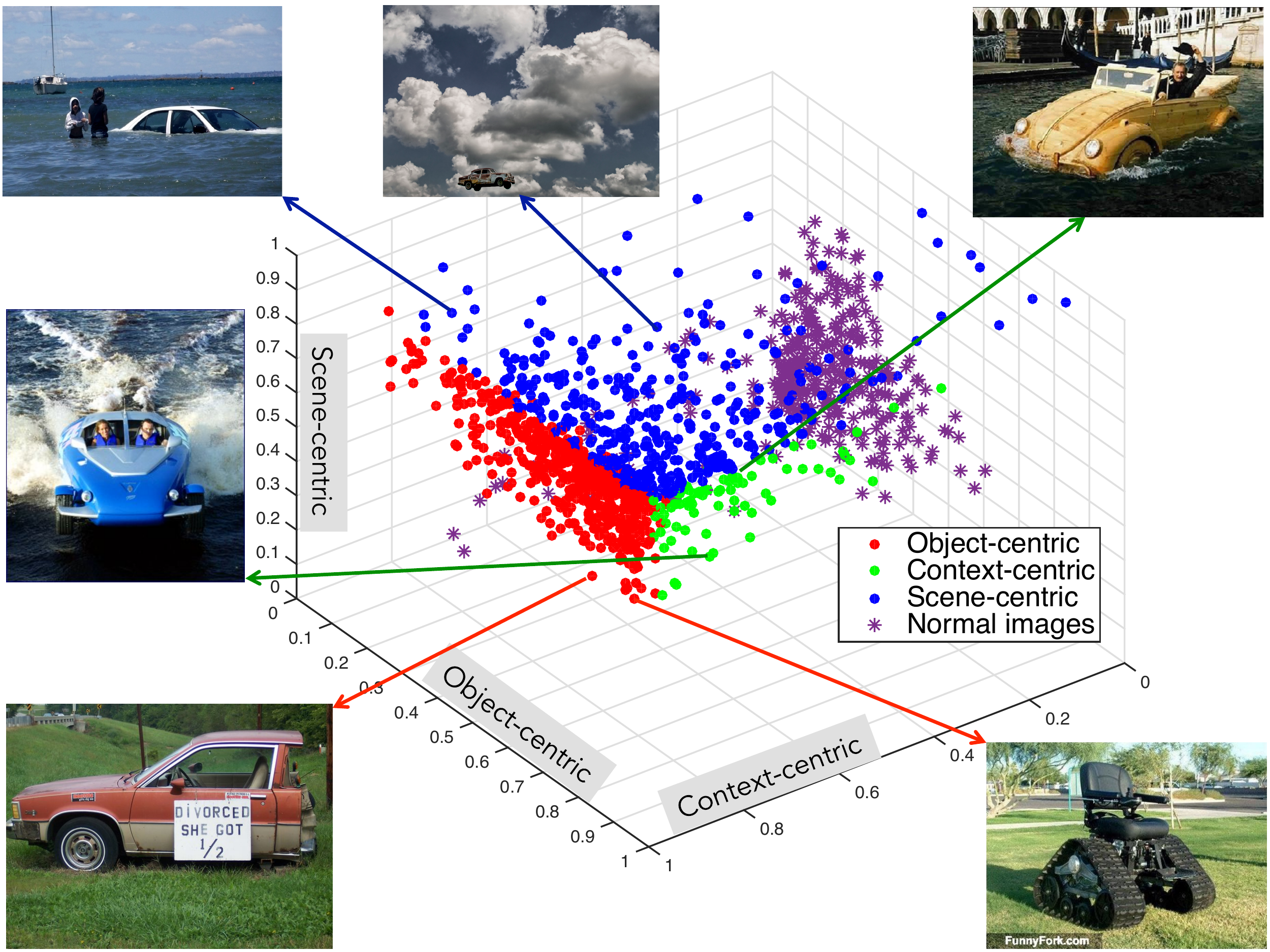}
		\caption{Plotting images in 3-D based on three surprise score that we get from our computational models. Points are colored based on the most important reason of abnormality in each image. Colorful clouds of abnormal images are separated from normal images (spread of purple stars close to the center of coordinate).}
		\vspace{-10pt}
		\label{fig:3Dplot}
	\end{figure}

	We use three reason-specific scores of abnormality to visualize abnormal images in a 3-D perspective. Figure~\ref{fig:3Dplot} shows this plot, where axis are surprise scores and data points are images, color coded based on the main reason of abnormality. For example, red dots are images that the most dominant reason of abnormality for them is object-centric. In this plot, we see how continuous surprise scores can spread abnormal images and we can find some boundaries between the main reasons of abnormality (three axes). More importantly, normal images (purple stars) are separable from abnormal images using these surprise scores. 
	
	In order to evaluate the quality of our model for reason prediction (abnormality reasoning), we compute the KL-divergence between three surprise scores of our model and the ground truth surprise scores for each image. We compute the ground truth scores by grouping and taking the average of response for 21 fine-grained reasons of abnormality in Turk experiment. We group these 21 fine-grained reasons based on the adopted taxonomy, and aggregate the corresponding responses to get three main surprise scores. We take the average of theses scores over all annotators for one image. The We measure the KL-divergence between scores of our final model and ground truth scores as 0.8142. Average human annotator predicts scores with KL-divergence of 0.6117. Interestingly, if we only use three raw surprise scores as the predicted scores, KL-divergence increases to 2.276. This verifies the value of parametric model for predicting more meaningful surprise scores, which are more similar to human judgments.
	
	In the last experiment, we classify abnormal images into three reasons of abnormality (abnormality reasoning) by picking the index of the reason that gives the highest surprise score. We compute the confusion matrix for this prediction as it is shown in Table~\ref{tab:conf}, where columns are ground truth labels and rows are the predicted labels.
	
	\begin{table}
		\begin{center}
			\begin{tabular}{|l|l|l|l|}
				\hline
				Object-centric abnormality &  301   &  62  &   41 \\ \hline
				Context-centric abnormality &   230   &  88  &   43 \\ \hline
				Scene-centric abnormality &   96   &  24  &   105  \\  
				\hline
			\end{tabular}
		\end{center}
		\caption{Confusion matrix for the task of abnormality reasoning. Rows are predicted labels and columns are ground truth given by the majority vote in the Turk experiment.}
		\vspace{-15pt}
		\label{tab:conf}
	\end{table}  
	
	\vspace{-5pt}
	\section{Conclusion \& Future Work}
	In this paper we approached the challenging research question: what make an image look abnormal? We made the biggest dataset of abnormal images and conducted a large-scale human subject experiment to investigate how humans think about abnormal images. We proposed a diverse list of abnormality reasons by human responses, and inferred a taxonomy of visual cues that make an image abnormal. Based on three major components of this taxonomy we built computer vision models that can detect an abnormal images and reason about this decision in terms of three surprise scores. Digging deeper into the categories of the inferred taxonomy is an important future work. 
	
	\paragraph{Acknowledgment:} This research was supported by NSF award IIS-1218872.
	\vspace{-5pt}
{	
\bibliographystyle{aaai}
\bibliography{camera_ready_one_page}}  
	
\end{document}